\documentclass[aps,prx,reprint,superscriptaddress,floatfix]{revtex4-2}

\usepackage{graphicx}
\usepackage{amsmath,amssymb}
\usepackage{booktabs}
\usepackage{siunitx}
\usepackage{hyperref}
\usepackage{xcolor}
\usepackage{enumitem}

\newcommand{\Dt}{D(t)}
\newcommand{\smax}{s_{\max}}
\newcommand{\Nzero}{N0}
\newcommand{\None}{N1}
\newcommand{\Ntwo}{N2}

\begin{document}

\title{Finite-Size Gradient Transport in Large Language Model Pretraining:
From Cascade Size to Intensive Transport Efficiency}

\author{Ping Wang}
\affiliation{Institute of High Energy Physics, Chinese Academy of Sciences, Beijing 100049, China}
\author{Yan-Qi Du}
\affiliation{Beijing Shenzhou Aerospace Software Technology Co., Ltd., Beijing 100094, China}

\date{\today}

\begin{abstract}
We introduce a finite-size gradient-transport framework for real language-model
training, based on five observables
$(D,z,\beta,\delta,v_{\mathrm{rel}})$: the cascade-size exponent $D$, the
duration exponent $z$, absolute and intensive transport exponents
$(\beta,\delta)$, and the relative transport efficiency $v_{\mathrm{rel}}$.
We analyze direct raw-gradient measurements from Pico-LM across four scales and
125 aligned steps, together with a five-scale Pythia companion dataset built
from 153 aligned checkpoint-difference update fields.
Using the same algebraic closure in both families, we find that Pico-LM and
Pythia share a near-unity cascade-size backbone but occupy distinct transport
regimes. Pico-LM shows positive duration scaling and negative intensive
intensive-efficiency scaling, whereas Pythia remains in a narrow band near the $D=1$
baseline with only weak positive efficiency scale dependence. Randomized-field
controls give nearly matched null floors in the intensive and duration channels,
so the contrast is best read as different real departures from a shared null
skeleton rather than as different null calibrations.
A further distinction appears in stepwise power-law compressibility: Pico-LM
retains clean duration and efficiency power laws, whereas Pythia preserves the
size backbone but shows weaker one-slope compressibility in the duration and
efficiency channels. We therefore treat compressibility as an observable of
internal transport organization rather than as a fit-quality footnote. The
association with external performance is correspondingly channel-level: the
strongest associations are carried by $v_{\mathrm{rel}}$ and normalized cascade
duration, while $D(t)$ acts as a shared size backbone without forming a
significant exponent-level performance association in either family. These results support a reusable
transport measurement framework for comparing real language-model training regimes without
claiming a universal fixed point or a first-principles derivation of neural
scaling laws.
\end{abstract}

\maketitle

\section{Introduction}
Neural scaling laws establish that large-language-model (LLM) pretraining loss follows robust
power-law dependencies on model scale, dataset size, and
compute~\cite{kaplan2020scaling,hoffmann2022chinchilla}. These regularities show that
large-scale training exhibits reproducible external scale dependence, but they do
not by themselves reveal how the internal organization of the gradient field evolves along
the training trajectory. In particular, loss and evaluation scaling do not provide a
scale-resolved measurement of how gradient-field organization changes across model
size and training time. This motivates a field-level, scale-resolved probe that
can be applied to aligned training snapshots, compared across model scales, and
calibrated against randomized null fields.

Long before modern LLM scaling-law studies, exact analyses of deep linear networks already exhibited
stage-like learning dynamics with plateaus and rapid transitions even in simplified
settings~\cite{saxe2013nonlinear}. Critical and near-critical signal-propagation analyses have similarly identified
order-to-chaos boundaries and trainable edge regimes as organizing principles for deep
networks~\cite{poole2016expressivity,schoenholz2017deep}.
Grokking, the delayed onset of generalization well after the model has fit the training set~\cite{power2022grokking},
shows that behavioral generalization can undergo transition-like changes that are not
apparent from the training-loss curve alone, and mechanistic follow-up work shows that
apparently sharp behavioral transitions can be preceded by smoother internal progress
measures~\cite{nanda2023grokking}. Modern optimization studies likewise point to
distinct large-learning-rate and edge-of-stability regimes during training~\cite{cohen2021edge,lewkowycz2020catapult},
and transformer-specific analyses have identified abrupt internal circuit emergence,
such as induction heads in language models~\cite{olsson2022induction}. Although these
phenomena arise in different architectures and tasks, they share a common implication:
scalar performance curves can hide internal reorganizations that unfold over training
time and across scale. A useful measurement should therefore be sensitive to both
global scale dependence and training-step-resolved internal dynamics.

Statistical physics provides a natural source of such probes. Self-organized
criticality (SOC)~\cite{bak1987soc} and the Olami-Feder-Christensen (OFC)
avalanche model~\cite{ofc1992} motivate a theoretical language for collective
cascade behavior in driven dissipative systems. In classical avalanche
ensembles, local load-transfer rules can sustain states where cascade sizes and
durations exhibit scale-invariant distributions with distribution
exponents~\cite{sethna2001crackling}. Finite-size scaling (FSS) addresses a
distinct but related question: how characteristic cascade scales change as the
system size is varied, with finite system size cutting off the growing correlation
scale near criticality and allowing size-dependent observables to be fit or
collapsed~\cite{fisher1972fss,sethna2001crackling}. We use these ideas as a
response-measurement language for gradient fields, not as a claim that LLM
pretraining realizes a classical SOC mechanism or a full crackling-noise
universality class~\cite{sethna2001crackling}. In the present setting, an avalanche is
not assumed to be a physical failure event, but is defined as the response of a
thresholded gradient field under a controlled redistribution rule. In this
finite-size transport sense, a characteristic cascade scale
$\smax \sim N^{D}$ encodes the extensivity of transport across system sizes $N$,
while the duration scaling $n_{\mathrm{steps}} \sim N^{z}$ separates spatial
reach from temporal persistence; these are not the avalanche-ensemble
distribution exponents of $P(S)$ or $P(T)$.

In two companion TDU-OFC studies~\cite{wang2025tduofc_a,wang2025tduofc_b}, we developed
the Thresholded Diffusion Update--OFC (TDU-OFC) framework as an offline, field-level
response probe of gradient-field reorganization. At each checkpoint, the
per-parameter gradient field is held fixed as a load field for controlled
threshold-redistribution cascades on an OFC-inspired probe graph. The resulting
cascade response is characterized by $\smax(N,t)$, and cross-scale fitting yields
the finite-size cascade-size exponent $D(t)$. Applied to controlled grokking
benchmarks, including XOR in MLPs~\cite{wang2025tduofc_a} and modular addition in
Transformers~\cite{wang2025tduofc_b}, this $D$-centered analysis revealed a
localized crossing of the $D=1$ baseline near the generalization transition, with
a task-dependent crossing direction.

Those studies also used synthetic gradient ensembles to test the dependence of
the probe on the underlying redistribution graph, finding graph-insensitive
finite-size scaling in the controlled settings analyzed~\cite{wang2025tduofc_a}.
Together, they established $D(t)$, extracted through the TDU-OFC avalanche
response, as a scale-resolved observable of gradient-field geometry in controlled
toy settings. However, they remained restricted to controlled toy tasks and were
centered primarily on a single size exponent. The present work asks whether the
same response-measurement language can resolve the internal organization of real
language-model pretraining trajectories.

The next step is to move this physics-inspired measurement program from
controlled grokking benchmarks to direct measurements of real language-model
pretraining trajectories.
A single crossing statistic, however, no longer suffices in this setting. In both
model families analyzed below, the size exponent $D(t)$ remains close to the
$D=1$ baseline through most of the post-startup trajectory, so the localized
$D=1$ crossings that organized the toy-grokking analyses~\cite{wang2025tduofc_a,wang2025tduofc_b}
are absent here as sharp organizing markers. Real systems can share this
near-unity size backbone while differing substantially in cascade duration,
per-parameter transport efficiency, and the degree to which their internal
channels remain compressible into clean power laws at each training step. The
central question is therefore no longer whether different families realize the
same single-exponent crossing pattern, but whether a common transport language
can resolve distinct real-training regimes.

Building on the TDU-OFC cascade response probe~\cite{wang2025tduofc_a,wang2025tduofc_b},
we therefore extend the analysis from a single size exponent $D(t)$ to a
five-quantity finite-size gradient-transport characterization. In this terminology,
finite-size gradient transport names the measurement framework, TDU-OFC names the
concrete gradient-response avalanche probe, and the multi-channel transport
observables are the fitted and directly measured readouts. In addition to the size
exponent $D$, we track the duration exponent $z$, the absolute transport exponent
$\beta$, the intensive transport exponent $\delta$, and the time-resolved efficiency
observable
$v_{\mathrm{rel}}=\smax/(N\,n_{\mathrm{steps}})$. These observables are linked by
algebraic closure relations because transport per step and transport per parameter
are constructed from $\smax$, $N$, and $n_{\mathrm{steps}}$. Together, they separate
three physically distinct aspects of cascade organization: the spatial extensivity
of transport, encoded in $D$; the temporal persistence of a cascade relative to its
spatial reach, encoded in $z$; and the intensive per-parameter transport efficiency,
encoded in $\delta$ and $v_{\mathrm{rel}}$.

Two complementary empirical families ground this framework. Pico-LM~\cite{pico2025}
serves as the raw-gradient reference family: it provides four model scales,
125 aligned training steps, and direct gradient snapshots from which TDU-OFC 
cascade response probe can extract complete per-step transport outputs, including
$n_{\mathrm{steps}}(t)$ at every aligned step. From the broader
Pythia/PolyPythias releases~\cite{biderman2023pythia,vanderwal2025polypythias},
which include larger model scales, we use five sub-1B scales (14M--410M) as a
companion family in which the analyzed field is constructed from
checkpoint-to-checkpoint parameter differences. This
range partially overlaps the Pico-LM parameter range while extending the analysis
to a distinct field representation. We therefore do not treat the two families as
numerically interchangeable measurements of the same microscopic field. Instead,
we pair them to ask whether the same transport framework resolves coherent
within-family regimes and interpretable cross-family contrasts under a shared
analysis protocol.

The paper makes three contributions. First, the five-quantity description
$(D,z,\beta,\delta,v_{\mathrm{rel}})$ provides a compact, algebraically
self-consistent transport characterization of real training, combining fitted
finite-size exponents with the directly measured intensive observable
$v_{\mathrm{rel}}$. Second, the two families share a near-unity size backbone
but occupy distinct transport regimes: the contrast appears most clearly in
$v_{\mathrm{rel}}$, is resolved quantitatively by the duration/intensive
decomposition, and is further clarified by stepwise compressibility, namely the
degree to which a transport channel admits a clean cross-scale power-law slope at
each training step. Third, the cross-family comparison can be placed on more
systematic footing than a visual contrast alone: randomized null baselines
provide a nearly shared reference across families, while the connection to
external scaling is carried primarily by the transport-efficiency channels rather
than by the size exponent $D(t)$ itself.

Taken together with the prior toy-grokking studies that introduced the TDU-OFC cascade response
probe and the single-exponent $D(t)$ readout~\cite{wang2025tduofc_a,wang2025tduofc_b},
the present work carries the same offline transport measurement from controlled
grokking benchmarks into real LLM pretraining. The resulting progression is a
controlled minimal demonstration in XOR/MLP, a cross-task and cross-architecture
extension in modular-addition Transformers, and here a real-pretraining extension
across LLM families. Across that sequence, the effective cascade dimension
$D(t)$ changes role: it acts as a transition-localized marker in controlled
grokking, but becomes a backbone coordinate for real-pretraining transport,
where the single-index picture must be extended into a multi-channel
finite-size gradient-transport description.

The exchange between physics and AI here is two-way. Tools developed for
non-equilibrium statistical mechanics, including FSS, OFC-style avalanche
analysis, and randomized null hierarchies, supply a field-level,
scale-resolved language for describing pretraining dynamics. Conversely, real
LLM pretraining provides a data-rich non-equilibrium setting in which the
relevant organization is not exhausted by a single scaling exponent. This
setting forces us to treat stepwise compressibility---the degree to which a
transport channel admits a single cross-scale power-law slope at each training
step---as a transport observable in its own right, rather than as a fit-quality
footnote. Although this observable is not usually treated as a primary observable
in classical SOC studies, the LLM setting suggests that it may be useful more
broadly for non-equilibrium scaling analyses in systems with evolving internal
transport channels.

Our conclusions are deliberately limited in scope. We do not claim a universal fixed
point, a universal $D(t)$-crossing pattern shared across architectures, that
$D\approx1$ by itself establishes criticality, or a direct microscopic derivation
of neural scaling laws. Instead, we argue that real pretraining is better
described by a multi-observable transport structure than by any single crossing
statistic. The paper proceeds accordingly: it first establishes the primary
$v_{\mathrm{rel}}$ contrast and the five-quantity closure, then shows how the
time-resolved decomposition, compressibility contrast, and shared null skeleton
make that contrast interpretable, and finally discusses a conservative
channel-level association between internal transport and external performance
scaling, accompanied by an exponent-level check.

\section{Methods}
\subsection{TDU-OFC cascade method}
\label{sec:tduofc}
We use the same offline TDU-OFC
avalanche probe introduced in our prior toy-grokking
studies~\cite{wang2025tduofc_a,wang2025tduofc_b}. The present work applies
this probe to saved field snapshots from real language-model pretraining
trajectories, including raw-gradient fields and checkpoint-difference update
fields, and extends the readout from a single size exponent to the
five-quantity transport characterization defined in Sec.~\ref{sec:fss_window}.
TDU-OFC is used here as a diagnostic response probe for gradient and update fields: it measures the
relaxation response of a thresholded field under a controlled redistribution
rule, rather than serving as a literal mechanistic model of pretraining
dynamics.
Starting from the observed signed field at one training step, the probe
identifies active nodes by field magnitude and then relaxes the signed field
on a fixed redistribution graph under an OFC-style threshold-relaxation rule.
Because gradients and update fields are directional quantities, the probe uses
field magnitude only to select large local update signals and preserves sign
during redistribution.
Localized cascades indicate localized transport under the probe, whereas
extended cascades indicate broader threshold-mediated transport within the
field snapshot. TDU-OFC cascade response probe therefore provides a compact transport
characterization of the field that can be compared systematically across
training steps and model scales.

Operationally, TDU-OFC is an offline avalanche diagnostic that extracts
cascade statistics from field snapshots through an OFC-style threshold
redistribution rule~\cite{ofc1992,wang2025tduofc_a,wang2025tduofc_b}.
For a model with $N$ trainable parameters, we construct a fixed Barab\'{a}si--Albert (BA) graph~\cite{barabasi1999}
with attachment parameter $m=2$ over the $N$ parameter nodes. The BA graph
provides a generic sparse redistribution graph and is not intended to encode
the architectural connectivity of the neural network. All reported runs use
topology seed 42; for each field size $N$, the resulting graph is cached by
$(N,m,\mathrm{seed})$ and reused across aligned snapshots with that field size.
Control experiments with
synthetic i.i.d.\ Gaussian fields found that the extracted $D$ is insensitive
to the tested graph-topology choices (coefficient of variation
$<0.4\%$)~\cite{wang2025tduofc_a}. This supports the interpretation that, in
the analyzed settings, deviations from the null are dominated by the organization
of the input field rather than by the chosen probe graph.

Given a saved field snapshot $u(t)\in\mathbb{R}^{N}$, with $u_i(t)=g_i(t)$
for raw-gradient measurements and $u_i(t)=\Delta\theta_i(t)$ for
checkpoint-difference measurements, we set a fixed threshold
\begin{equation}
  \tau(t)=Q_{90}\bigl(\{|u_i(t)|\}_{i=1}^{N}\bigr),
\end{equation}
the empirical 90th percentile of the initial field magnitude. At relaxation
step $\ell$, the active set is recomputed as
\begin{equation}
  A_\ell(t)=\{i: |u_i^{(\ell)}(t)|>\tau(t)\},
\end{equation}
with $\tau(t)$ held fixed throughout the relaxation of that snapshot. All
active nodes are updated synchronously, with incoming contributions accumulated
from the previous relaxation state. Equivalently, for each node $i$, with
$k_r$ denoting the degree of source node $r$ in the redistribution graph and
$r\sim i$ denoting adjacency in that graph,
\begin{equation}
  u_i^{(\ell+1)} =
  \left(1-\alpha\,\mathbf{1}_{i\in A_\ell}\right)u_i^{(\ell)}
  + \sum_{\substack{r\in A_\ell\\ r\sim i}}
    \frac{\alpha\,u_r^{(\ell)}}{k_r}.
  \label{eq:ofc_rule}
\end{equation}
The first correction decays active source nodes, while the sum accumulates
signed contributions from active neighbors.
The redistribution fraction $\alpha$ controls cascade intensity. The degree
normalization in Eq.~\eqref{eq:ofc_rule} conserves the signed redistributed
field at each synchronous relaxation step. Because activity is defined by
magnitude whereas redistribution preserves sign, the magnitude field can
change through constructive or destructive interference. In this signed-field
sense, the probe is quasi-conservative and analogous to the load-transfer
structure of OFC-type models~\cite{ofc1992}; the resulting cascade extent
reflects signed field organization under the threshold-response rule rather
than unbounded external driving. Relaxation is iterated until no active nodes
remain or until a ceiling of 500 relaxation steps is reached. Ceiling-limited
snapshots are recorded with $n_{\mathrm{steps}}=500$; when constructing
windowed summaries and applying the degeneracy criterion described below, these
snapshots are excluded from the corresponding summary statistics.
All computations reported here use $\alpha=0.3$; robustness to $\alpha$ and
threshold percentile was reported in Ref.~\cite{wang2025tduofc_a}.
For computational efficiency, when the field size exceeds $10^7$ entries the
empirical 90th percentile of the field magnitude is estimated from a
uniform-random subsample of $10^7$ magnitude values drawn once per snapshot;
this stays below the CUDA \texttt{torch.quantile} input-size limit of
approximately $2^{24}$ elements. In validation checks
against full-quantile calculations on smaller fields, this subsampling
procedure gave
$|Q_{90}^{\mathrm{subsample}}-Q_{90}^{\mathrm{full}}|/Q_{90}^{\mathrm{full}}<0.1\%$.

The cascade duration $n_{\mathrm{steps}}$ is the number of relaxation steps
until termination, and the cascade size
\begin{equation}
  s(t)=\sum_{\ell=0}^{n_{\mathrm{steps}}-1}|A_\ell(t)|
\end{equation}
is the total count of active update events accumulated over all relaxation
steps. For the one-relaxation-per-snapshot protocol used here, $s(t)$ is the
snapshot-level characteristic size entering the finite-size analysis and is
denoted $\smax(N,t)$ in Sec.~\ref{sec:fss_window}. Both quantities are computed from
field copies that do not enter the parameter update, so the cascade computation
leaves the training trajectory unperturbed and is therefore fully offline.

\subsection{Finite-size scaling exponents and transport observables}
\label{sec:fss_window}
At each training step $t$ and for each model scale $N$, the TDU-OFC method (Sec.~\ref{sec:tduofc})
yields $s(N,t)$ and $n_{\mathrm{steps}}(N,t)$. For consistency with our earlier TDU-OFC notation, we denote
this snapshot-level cascade size by $\smax(N,t)$. In the present one-relaxation-per-snapshot
protocol, $\smax$ refers to the integrated size of that snapshot-driven cascade,
not to a maximum over an avalanche ensemble or to the peak number of active
nodes in a relaxation step. At fixed $t$, we fit
\begin{equation}
\log \smax = \Dt \log N + c
\end{equation}
across model scales to obtain $\Dt$ and fit quality $R^2$. All slopes are
estimated by ordinary least-squares regression in log--log space over the
common set of available model scales at that aligned step. Accordingly, $D$
should be read as a cascade-size scaling
exponent, more specifically an integrated transport-size exponent for one snapshot-driven
cascade, rather than as a literal geometric dimension. As an aid to
intuition, one may regard it as an abstract measure of the extensivity of cascade support
describing how the effective support of a single cascade grows with $N$, but not as a 
strictly geometric dimension. We additionally fit cascade duration
\begin{equation}
\log n_{\mathrm{steps}} = z(t)\log N + c_z
\end{equation}
and define
\begin{equation}
v_{\mathrm{abs}} = \frac{\smax}{n_{\mathrm{steps}}}, \qquad
v_{\mathrm{rel}} = \frac{\smax}{N\,n_{\mathrm{steps}}} = \frac{v_{\mathrm{abs}}}{N}.
\end{equation}
Writing
\begin{equation}
v_{\mathrm{abs}} \sim N^\beta, \qquad v_{\mathrm{rel}} \sim N^\delta,
\end{equation}
the log-linear transport framework closes as
\begin{equation}
\beta = D-z, \qquad \delta = \beta - 1 = D-z-1.
\end{equation}
These equalities serve as cross-checks within the transport framework. They
follow algebraically when the same model scales and log-linear fitting
procedure are used, since
$\log v_{\mathrm{abs}}=\log\smax-\log n_{\mathrm{steps}}$ and
$\log v_{\mathrm{rel}}=\log\smax-\log n_{\mathrm{steps}}-\log N$.
They also make the transport interpretation explicit: $D-1$ captures the scale
dependence of $\smax/N$ (i.e., $N^{D-1}$), while $z$ captures duration scaling
exponent, so the finite-size scaling of $v_{\mathrm{rel}}$ is organized by
their difference $\delta=D-1-z$.
These exponents are finite-size transport exponents measured across model
scale at fixed training step, not avalanche-ensemble distribution exponents.
We therefore do not test, or require, the classical crackling-noise scaling
relation between avalanche-size, duration, and conditional size-duration
distribution exponents.

Because $D(t)$, $z(t)$, and $\delta(t)$ are estimated from per-step cross-scale fits,
regime-level summaries are reported only over a temporal interval in which the transport
observables fluctuate within an approximately stationary band. Startup transients are excluded
because they reflect rapid initialization dynamics rather than the characteristic training
regime. Operationally, we identify the stable window from the per-model
$v_{\mathrm{rel}}(t)$ trajectories: the lower boundary is the first aligned step after the
startup relaxation has ended, and the upper boundary is the last aligned step before
systematic late-time drift becomes visible. The resulting window is fixed at
the family level and applied uniformly to all transport observables. This gives
steps $5\,000$--$60\,000$ for Pico-LM and steps $4\,000$--$73\,000$ for Pythia,
corresponding to 56 and 70 common aligned steps, respectively. For Pythia,
snapshots failing the standard degeneracy criterion, namely $\smax=0$ (no
cascade event formed) or $n_{\mathrm{steps}}=n_{\mathrm{steps}}^{\max}=500$ (a
ceiling-limited cascade that hit the relaxation iteration cap), are excluded
before any windowed averaging. On the
five-scale common grid this removes the four earliest log-spaced checkpoints
(steps $1,2,4,8$): step $2$ is ceiling-limited
($n_{\mathrm{steps}}=n_{\mathrm{steps}}^{\max}$) across all model scales, while
steps $1$, $4$, and $8$ fail the $\smax>0$ criterion in at least one scale at
each step. In the affected scales these snapshots are dominated by
initialization noise rather than the field organization of interest, and the
common-grid requirement that all five scales pass the degeneracy filter
therefore excludes these four checkpoints from the analyzed window.
Qualitative regime assignments are unchanged under small shifts of these boundaries.
Within the stable window, we retain all per-step exponent estimates without additional
$R^2$-based filtering, because for weak-signal channels such as $z$ and $\delta$ such
filtering would preferentially retain larger apparent slopes and bias the estimates away
from zero.

\subsection{Data representation and comparability}
The two training families differ in the gradient-field representation supplied to TDU-OFC cascade response probe.
Pico-LM provides direct access to the instantaneous gradient vector
$g_t = \nabla_\theta \mathcal{L}_t \in \mathbb{R}^N$. Pythia provides only released weight
checkpoints; we therefore analyze the accumulated parameter-update proxy
\begin{equation}
\Delta \theta_t = \theta_{t+\Delta t}-\theta_t,
\end{equation}
defined as the difference between adjacent released checkpoints. In the released
Pythia setup~\cite{biderman2023pythia}, training uses Adam~\cite{kingma2014adam} with
$\beta_1=0.9$, $\beta_2=0.95$, cosine learning-rate decay over 143k steps, and 1\% warmup;
checkpoints are released at log-spaced steps $\{1,2,4,\ldots,512\}$ during the early
training phase and at uniform 1000-step intervals thereafter (steps $1000,2000,\ldots,143000$).
The analysis window adopted in this paper (Pythia steps $4\,000$--$73\,000$, see
Sec.~\ref{sec:fss_window}) lies entirely in the uniform-spacing regime, so every $\Delta\theta_t$
entering the present transport analysis corresponds to a fixed checkpoint spacing
$\Delta t = 1000$ iterations. Within the analyzed window, the learning-rate
change over a 1000-step interval remains at the percent level and below roughly
3\%, so each $\Delta\theta_t$ is well-approximated as a short-window
accumulated update field rather than a single-step gradient snapshot.

This proxy should not be identified with a raw gradient. Because Adam applies parameterwise
preconditioning, $\Delta\theta_t$ is best interpreted as an optimizer-mediated update field.
This distinction does not preclude the present within-family transport analysis. TDU-OFC thresholds
each field at its own empirical 90th percentile, so positive global scalar
rescalings leave the active-set sequence and cascade observables invariant;
differences in learning-rate amplitude across model sizes therefore do not by
themselves generate artificial cross-scale transport trends, while
optimizer-mediated changes in field shape remain part of the update-field
representation being measured. Moreover, with Adam
$(\beta_1=0.9,\beta_2=0.95)$ and 1000-step checkpoint spacing, each checkpoint difference aggregates
many locally preconditioned updates and is best read as a short-window time-averaged update field.
We therefore treat Pythia as a companion update-field family rather than as a second raw-gradient
measurement.

Accordingly, our strongest quantitative claims are made within a single model family under a fixed
field representation. Cross-family comparisons are limited to regime-level contrasts and to the
structural validity of the five-exponent closure. Pico and Pythia should therefore be viewed as
independent tests of the same transport framework, not as numerically interchangeable measurements
of a single microscopic field. Because the same null constructions are applied to both families,
their calibration floors can still be compared under the same null protocol without erasing the representational distinction.

\subsection{Null decomposition}
To separate null calibration, marginal-distribution effects, and assignment
effects, we compute four matched trajectories for a given observable $X$:
\begin{itemize}[leftmargin=*,itemsep=1pt,topsep=2pt]
    \item $X_{\mathrm{real}}$: the original signed field $u(t)$;
    \item $X_{\Nzero}$: an i.i.d.\ standard Gaussian field;
    \item $X_{\None}$: an i.i.d.\ Gaussian field whose mean and variance are
    matched to the signed entries of the real field at the same checkpoint;
    \item $X_{\Ntwo}$: a signed-value-preserving permutation null, obtained by
    randomly permuting the entries of the real field over the fixed probe-graph nodes.
\end{itemize}
The near-identical cascade statistics of $\Nzero$ and $\None$ indicate that
the Gaussian baseline is insensitive to global field-scale variation under the
percentile-thresholded probe. The $\Ntwo$ null preserves the empirical signed
marginal distribution while removing the original assignment of values to
parameter/probe-graph nodes.
We then decompose
\begin{align}
    \Delta X_{\mathrm{total}} &= X_{\mathrm{real}}-X_{\Nzero},\\
    \Delta X_{\mathrm{dist}} &= X_{\Ntwo}-X_{\Nzero},\\
    \Delta X_{\mathrm{assign}} &= X_{\mathrm{real}}-X_{\Ntwo}.
\end{align}
By construction,
\begin{equation}
    \Delta X_{\mathrm{total}}
    = \Delta X_{\mathrm{dist}} + \Delta X_{\mathrm{assign}}.
\end{equation}
Our null hierarchy supports a two-stage interpretation. The transition
$\Nzero/\None \rightarrow \Ntwo$ captures departures from the Gaussian
reference caused by the empirical signed marginal distribution, whereas
$\Ntwo \rightarrow \mathrm{real}$ captures the residual contribution of the
original value-to-node assignment. Thus, when $\Ntwo$ lies close to the
real-data curve for a given observable, the conservative reading is that
assignment effects detectable by the fixed probe construction are small for
that observable. We do not take this as evidence for graph-family
insensitivity, since graph family is not varied in the present study.

In the Results, this hierarchy is used in two distinct ways. The main cross-family null
comparison is reported in the duration channel $z$, where it defines the shared null floor
and the null-subtracted contrast shown in Fig.~\ref{fig:shared_null_skeleton}. The same
hierarchy is also applied to the size exponent $D$, but there it is used only for
family-specific mechanism readings rather than as a primary cross-family figure.

At each checkpoint, the null baselines $\Nzero/\None/\Ntwo$ are generated once
using a deterministic checkpoint-specific seed, with base seed~42 shifted by
the released training-step value $t$, i.e.\ $\mathrm{seed}=42+t$. Here $t$
takes the released checkpoint step rather than a $0,1,2,\ldots$ enumeration
index. The three nulls should therefore be interpreted as matched
single-realization controls, not as independent Monte Carlo estimates. No
averaging over multiple null seeds is performed. Consequently, per-step null
estimates retain a single-realization noise component, while the stable-window
summaries average over 56 snapshots (Pico-LM) and 70 snapshots (Pythia) and
therefore reduce, but do not eliminate, null-seed noise.

The pure Gaussian null $\Nzero$ is special in this respect. Since $\Nzero$ is independent of the
real field and depends only on $(N,\mathrm{graph},\mathrm{seed})$ within a fixed model, the
checkpoint sequence provides a practical sensitivity check to single-seed
Gaussian fluctuations, with each checkpoint contributing a fresh Gaussian
realization. This does not replace a true multi-seed Monte Carlo average, but
it helps verify that the $\Nzero$ floor is not dominated by one particular
Gaussian draw. This strengthening applies only to $\Nzero$; by contrast,
$\None$ and $\Ntwo$ remain checkpoint-dependent controls because they inherit
moment or marginal information from the real field at each checkpoint.

\subsection{Datasets and evidence tiers}
\subsubsection{Pico-LM (primary raw-gradient family)}
Pico-LM~\cite{pico2025} is the primary raw-gradient family analyzed in this study.
We use the four released \texttt{pico-decoder} models (tiny, small, medium, and large),
a LLaMA-style decoder-only family trained under matched settings on
\texttt{pretokenized-dolma}, derived from Dolma~\cite{soldaini2024dolma}.
These runs have nominal model sizes of 11M, 65M, 181M, and 570M parameters and are
available at 125 aligned checkpoints from step 1k through step 125k. The released
artifacts include direct raw-gradient snapshots at every checkpoint.

The released snapshots cover the same tensor types at every scale: the attention value
and output projections (\texttt{v\_proj} and \texttt{o\_proj}) and the MLP down-projection
(\texttt{swiglu.w\_2}) across the 12 released transformer blocks in each model,
for a total of 36 tensors.
The resulting analyzed gradient-field sizes are approximately $0.59$M, $9.44$M,
$37.75$M, and $150.99$M elements for tiny, small, medium, and large, respectively.
These partial-gradient element counts, rather than the nominal total-parameter counts,
therefore define the scale variable $N$ for all Pico-LM FSS fits. This partial coverage
is a constraint of the public Pico release, so the present analysis characterizes cascade
dynamics in the attention output and MLP down-projection channels of the gradient field;
the effect of the excluded channels is not assessed here.

Applying TDU-OFC cascade response probe to these released gradients yields the full set of per-checkpoint
cascade observables, including $n_{\mathrm{steps}}(t)$, at all four scales and all
125 aligned checkpoints. For comparisons to external evaluation performance, we use the per-checkpoint
perplexity values distributed with the Pico artifacts on the \texttt{pico-paloma-tinsy}
subset of Paloma~\cite{pico2025,magnusson2023paloma}.

\subsubsection{Pythia/PolyPythias (checkpoint-difference companion family)}
The checkpoint-difference companion family is drawn from the Pythia project~\cite{biderman2023pythia}
and the PolyPythias seed extensions~\cite{vanderwal2025polypythias}. These are
GPT-NeoX-based autoregressive language models~\cite{gpt-neox-library} trained on the
Pile~\cite{gao2020pile}. In the present analysis we use the five seed-3 released runs
\texttt{pythia-14m-seed3}, \texttt{pythia-31m-seed3}, \texttt{pythia-70m-seed3},
\texttt{pythia-160m-seed3}, and \texttt{pythia-410m-seed3}, spanning nominal sizes from
14M to 410M and yielding 153 aligned checkpoint-difference fields on the shared
released step grid. Because these models are publicly released as checkpoints rather than raw-gradient
snapshots, the analyzed field is the checkpoint difference between aligned states, used here as a
proxy for the parameter-update field; trainable-parameter counts are computed explicitly for each
model, with persistent buffers excluded.

We restrict the companion dataset to these five sub-1B runs because they form a complete,
internally consistent size ladder for the present study and partially overlap the Pico-LM analyzed
size range. Their role is not to duplicate the Pico raw-gradient measurement, but to test the same
transport framework on a second model family under a different field representation.
Several properties of the checkpoint-difference proxy support its use here.
The checkpoint interval is fixed at 1000 training steps throughout the main run, 
so each $\Delta\theta$ field is a short-window accumulated update rather than an 
arbitrary-length mixture. Within any such 1000-step window the cosine learning-rate 
schedule changes by at most a few percent, so $\Delta\theta$ approximates a windowed 
time-averaged update field.
Adam's first- and second-moment estimates average over tens of steps for
$\beta_1=0.9$ and $\beta_2=0.95$, much shorter than the 1000-step checkpoint
interval, so each $\Delta\theta$ integrates many locally preconditioned updates
and carries a stable window-average meaning.
Finally, the TDU-OFC threshold is set relative to the field itself, making the
active-set sequence and cascade observables invariant under positive global
amplitude rescaling.
Pico and Pythia should therefore be read as independent tests of the same
transport-observability framework, not as two numerically interchangeable
measurements of an identical microscopic field; model-family-specific
quantitative claims (null decompositions, partial correlations, regime
exponents) remain strictly within-family, while cross-family comparisons are
restricted to regime-level structure.

\subsubsection{External performance metrics and learning-rate partial correlations}

For Pico-LM, the external performance metric is a perplexity scaling exponent
$\beta_{\mathrm{PPL}}(t)$, defined as the log-log slope
$\partial\log\mathrm{PPL}/\partial\log N$ fitted across the four training scales at
each step using the per-checkpoint Paloma perplexity~\cite{pico2025,magnusson2023paloma}
evaluated on the \texttt{pico-paloma-tinsy} subset.

For Pythia, the external performance metric is a zero-shot accuracy scaling exponent
$\alpha_{\mathrm{acc}}(t)$, defined as the log-log slope
$\partial\log\bar{a}/\partial\log N$ fitted across the 70M/160M/410M evaluation
line, where $\bar{a}(N,t)$ is the mean normalized accuracy over 65 common zero-shot
tasks, comprising 57 Massive Multitask Language Understanding (MMLU) subtasks~\cite{hendrycks2021mmlu}
and 8 additional standard zero-shot benchmarks implemented in the LM Evaluation
Harness~\cite{eval-harness}, with CrowS-Pairs social-bias tasks excluded.
For each task we use the released
normalized accuracy when available (\texttt{acc\_norm}) and otherwise accuracy
(\texttt{acc}); the log-slope fit uses a small positive floor only to avoid
taking logs of nonpositive values. The 14M and 31M seed-3 scales enter the
internal transport analysis but do not have matching released zero-shot
evaluation traces in this evaluation set, so the external accuracy slope is fit
on the 70M/160M/410M line.

Neither quantity is intended as an asymptotic Kaplan--Chinchilla exponent; both
are used here only as finite-scale summaries of how external scale separation
co-evolves with internal transport reorganization.

To check whether the reported transport--performance correlations are driven
primarily by the learning-rate schedule, we also report
``learning-rate-partial'' Pearson correlations computed after linearly
residualizing both variables against the reconstructed learning-rate schedule
$\eta(t)$ (linear-with-warmup for Pico-LM, as specified in the released
training configuration~\cite{pico2025}; cosine decay with linear warmup for
Pythia~\cite{biderman2023pythia}).

\subsection{Unified analysis conventions}
All reported Pico-LM and Pythia results are generated under a common analysis protocol.
At each aligned training step, finite-size fits are performed only on the model
scales that are available and pass the relevant inclusion criteria at that
step. Linear log-log exponents are reported only when at least three scales are
present; steps that do not meet this criterion are retained as per-scale
transport records but are excluded from FSS estimates and from any FSS-based
regime interpretation rather than extrapolated. 

The same fitting definitions,
inclusion rules, null constructions, and per-step summary statistics are then
applied to both families, so the contrasts reported below are not introduced by
family-specific changes in fitting or filtering procedure. This common protocol
standardizes the analysis, but it does not remove the representational
distinction between Pico-LM raw-gradient fields and Pythia
checkpoint-difference update fields discussed above.

\section{Results}
\begin{figure*}[!t]
    \centering
    \includegraphics[width=0.96\textwidth]{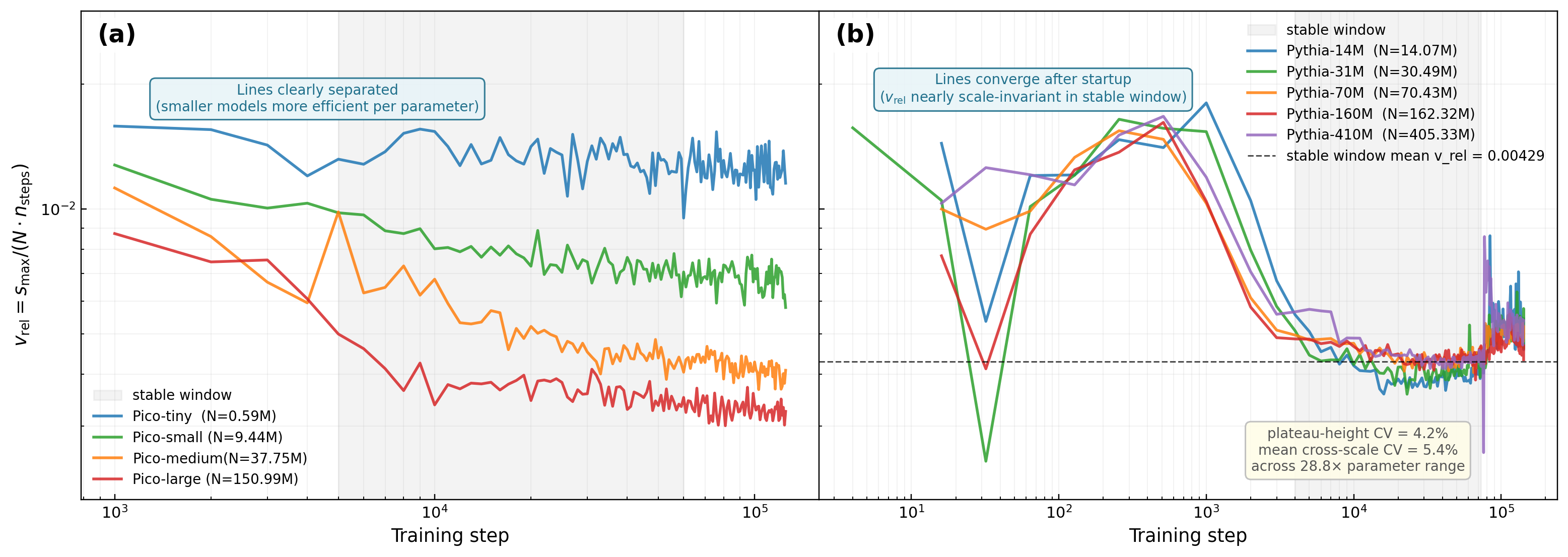}
    \caption[Intensive transport efficiency regime contrast]{
    Intensive transport efficiency, $v_{\mathrm{rel}}=\smax/(N\,n_{\mathrm{steps}})$, across the
    two model families.
    (a): Pico-LM. After the startup transient, the four-scale raw-gradient family retains strong scale
    ordering, with smaller models showing larger per-parameter, per-relaxation-step transport efficiency;
    the cross-model coefficient of variation of the stable-window means is $\approx 50.1\%$.
    (b): Pythia (14M/31M/70M/160M/410M). After the startup transient, the five-scale
    checkpoint-difference companion family approaches a narrow stable-window band with only
    weak residual positive scale dependence; the corresponding cross-model coefficient of variation
    is $\approx 4.2\%$.
    The figure shows the regime contrast directly in the intensive observable: persistent
    scale-sensitive transport in Pico-LM versus a narrow common-efficiency band across scales in Pythia.
    }
    \label{fig:vrel_hero}
\end{figure*}

\subsection{Intensive transport efficiency reveals a regime contrast}
Across the aligned steps retained for finite-size analysis, $\smax(N,t)$ is
well described by per-step power-law fits,
$\smax\sim N^{D(t)}$, in both model families. In the stable window, the size exponent
remains close to unity for both Pico-LM and Pythia
($D_{\mathrm{Pico}}=0.988\pm0.015$ and $D_{\mathrm{Pythia}}=0.996\pm0.009$; full
stable-window statistics in Fig.~\ref{fig:closure_table}), indicating a shared
near-unity size-scaling backbone. By itself, however, near-unity $D(t)$ does not
distinguish the two families.
A clearer cross-family contrast appears directly in the
intensive transport observable (Fig.~\ref{fig:vrel_hero}),
\begin{equation}
v_{\mathrm{rel}}(t)=\frac{\smax}{N\,n_{\mathrm{steps}}},
\end{equation}
which measures transport efficiency per parameter and per relaxation step.
In the Pythia checkpoint-difference companion family, startup transients give way to a narrow stable-window
band with low cross-scale variation (CV $=4.2\%$). In Pico-LM, by contrast,
$v_{\mathrm{rel}}$ remains strongly scale-ordered over the stable window, with a factor
of about $3.6$ between the smallest and largest models (CV $=50.1\%$).
The relevant contrast is therefore no longer a localized $D=1$ crossing, but
how differently the two families organize scale-normalized transport after
startup: Pythia approaches a narrow common-efficiency band, whereas Pico retains
persistent scale-sensitive transport. Because both families share near-unity
size scaling, the origin of this contrast must lie in the duration and
efficiency channels; the next subsection makes this decomposition explicit
through the five-quantity closure.

\subsection{The five-quantity closure resolves the regime contrast}
\begin{figure*}[!t]
    \centering
    \includegraphics[width=0.92\textwidth]{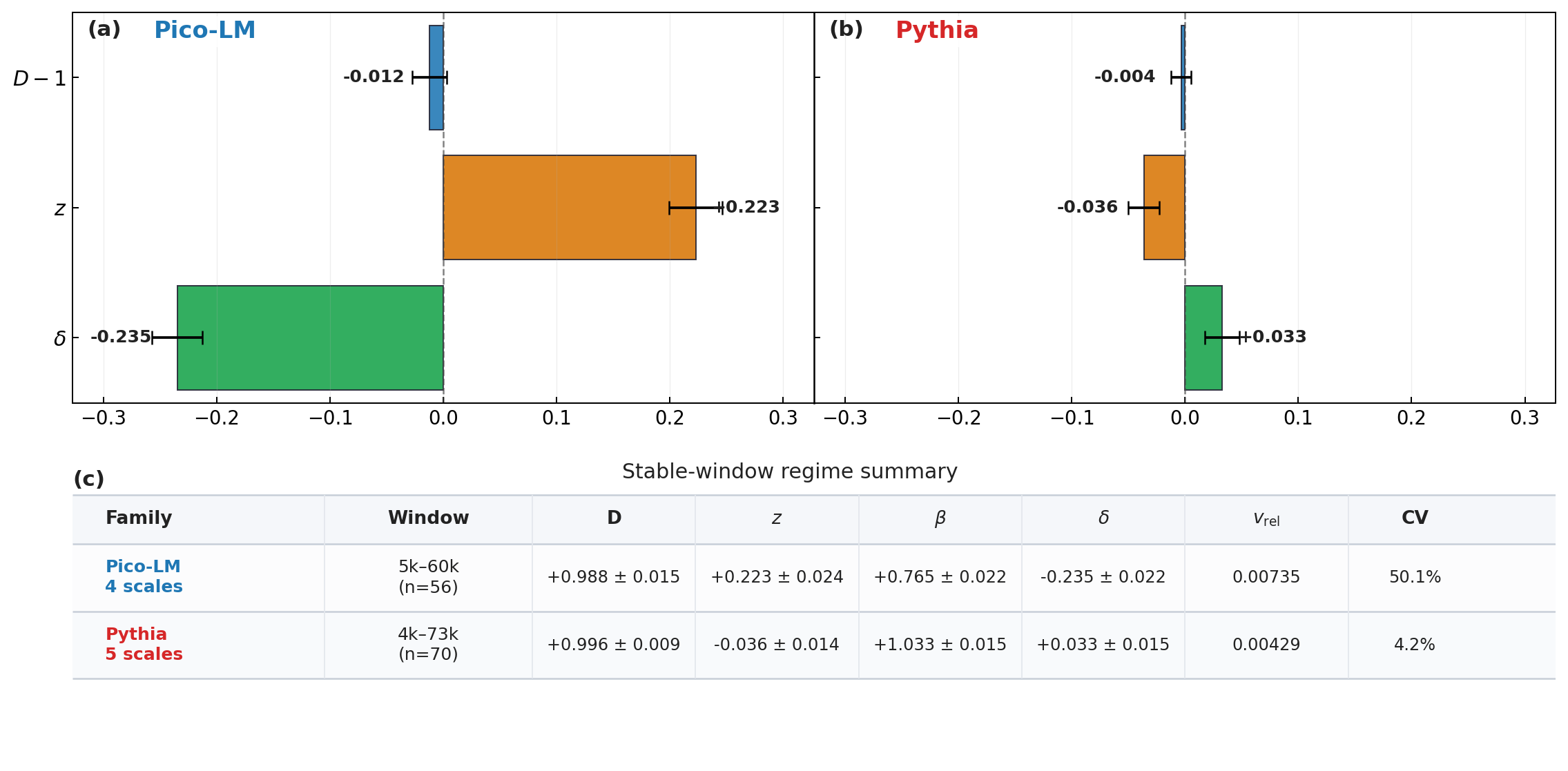}
    \caption[Five-quantity transport closure summary]{
    Five-quantity transport summary over the stable windows used in the main analysis.
    (a) and~(b) show the stable-window means of $D-1$, $z$, and $\delta$ for Pico-LM and
    Pythia, respectively, using the same vertical scale.
    (c) reports the corresponding stable-window means and standard deviations
    across stable-window steps for $D$, $z$, $\beta$, and $\delta$, together with
    the stable-window mean of $v_{\mathrm{rel}}$ and the cross-model coefficient of
    variation of the per-model $v_{\mathrm{rel}}$ plateau means.
    Because $\beta=D-z$ and $\delta=(D-1)-z$, the same algebraic closure relations
    hold for both families,
    but they partition near-unity size scaling differently.
    Pico-LM combines positive duration scaling with negative intensive transport scaling, whereas
    Pythia remains near zero in both duration and intensive-efficiency scaling, with only weak positive
    intensive-efficiency scaling ($\delta>0$).
    }
    \label{fig:closure_table}
\end{figure*}

The five-quantity closure identifies the channels responsible for the
$v_{\mathrm{rel}}$ contrast in Fig.~\ref{fig:vrel_hero}.
Figure~\ref{fig:closure_table} summarizes the stable-window transport signatures of the two families.
In Pico-LM, the stable window gives $D=0.988\pm0.015$, $z=+0.223\pm0.024$, and $\delta=-0.235\pm0.022$.
In Pythia, the corresponding five-scale stable window gives $D=0.996\pm0.009$, $z=-0.036\pm0.014$, and $\delta=+0.033\pm0.015$.
The directly fitted $\beta$ values agree with the closure-derived values,
$\beta=D-z\approx0.765$ for Pico-LM and $\beta\approx1.032$ for Pythia.
Because $\beta=D-z$ and $\delta=(D-1)-z=\beta-1$ under the same log-linear fitting definitions,
the closure relations hold algebraically in both families.
The closure therefore shows how a shared near-unity size backbone is partitioned between duration and intensive transport.
Pico combines positive duration scaling with negative intensive-efficiency scaling,
whereas Pythia remains near zero in both duration and intensive-efficiency scaling,
with weak negative $z$ and weak positive $\delta$.

\begin{figure*}[!t]
    \centering
    \includegraphics[width=0.96\textwidth]{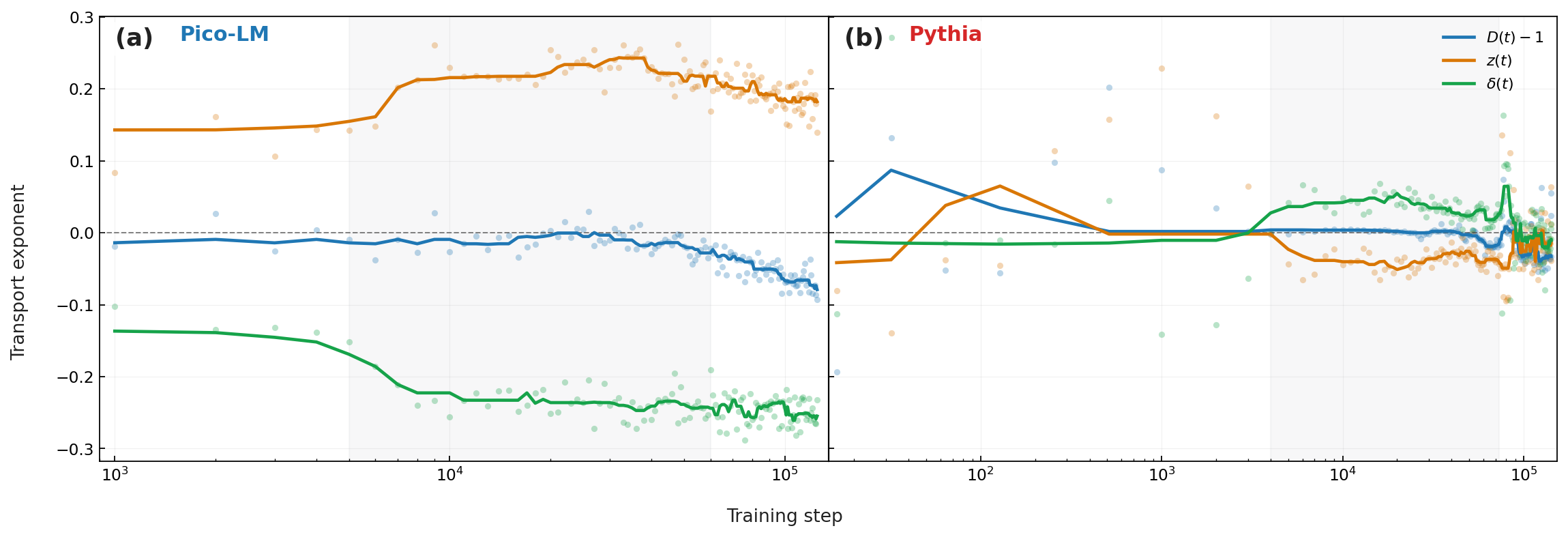}
    \caption[Time-resolved transport exponents]{
    Time-resolved transport exponents in the two model families.
    (a) and (b) show Pico-LM and Pythia, respectively, with $D(t)-1$, $z(t)$, and
    $\delta(t)=D(t)-1-z(t)$ plotted on the same vertical scale in both panels;
    stable windows are shaded.
    Markers show per-step exponent estimates, and solid curves show centered
    11-step rolling medians used only to guide the eye.
    All quoted stable-window statistics are computed from the unsmoothed per-step estimates,
    not from the rolling-median curves.
    In Pico-LM, $z(t)$ remains positive and $\delta(t)$ negative through the stable window.
    In Pythia, $D(t)-1$ stays close to zero while $z(t)$ remains weakly negative and
    $\delta(t)$ weakly positive over the stable window.
    The figure is the time-resolved counterpart of the stable-window regime summary in
    Fig.~\ref{fig:closure_table}.
    }
    \label{fig:transport_analysis}
\end{figure*}

Figure~\ref{fig:transport_analysis} provides the time-resolved counterpart to the stable-window summary.
In Pico-LM, $z(t)$ remains positive and $\delta(t)$ negative through the stable window and much of the
post-startup trajectory, showing that the stable-window summary reflects a broader post-startup pattern.
In Pythia, $D(t)-1$ stays close to zero while $z(t)$ and $\delta(t)$ remain small in magnitude but
retain opposite signs over the stable window, consistent with a close-to-unity band with weak positive
efficiency scale dependence rather than exact scale invariance ($\delta=0$) or the strong negative
efficiency scaling observed in Pico-LM.

This decomposition also clarifies what each evidence family supports.
Pico-LM serves as the raw-gradient reference family because it provides direct gradient-field
measurements. As summarized in Figs.~\ref{fig:closure_table} and \ref{fig:transport_analysis},
the stable-window interval, steps $5{,}000$--$60{,}000$, combines size scaling
close to the $D=1$ baseline with positive duration scaling and negative
intensive-efficiency scaling. Across the full run, $D(t)$ fluctuates around
unity without establishing a sustained above-baseline regime. The conservative
reading is therefore persistent negative efficiency scaling, rather than a
clean transition into a sustained $D=1$-locked or $D>1$ regime.

The five-scale Pythia line serves as the checkpoint-difference companion family. As
summarized in Figs.~\ref{fig:closure_table} and \ref{fig:transport_analysis}, the size
trajectory remains close to unity through most of the post-startup trajectory; over the stable
window, $D=0.996\pm0.009$, $z=-0.036\pm0.014$, and $\delta=+0.033\pm0.015$, while the
stable-window $v_{\mathrm{rel}}$ band shows only $4.2\%$ cross-scale variation. The appropriate
reading is therefore a close-to-unity band with weak positive efficiency scale dependence, not exact
transport invariance. The clearest cross-family null contrast appears in the duration channel,
where Fig.~\ref{fig:shared_null_skeleton} shows opposite-sign departures from a nearly shared
null floor.

\subsection{Stepwise compressibility as a transport-organization observable}
\begin{figure*}[!t]
    \centering
    \includegraphics[width=0.92\textwidth]{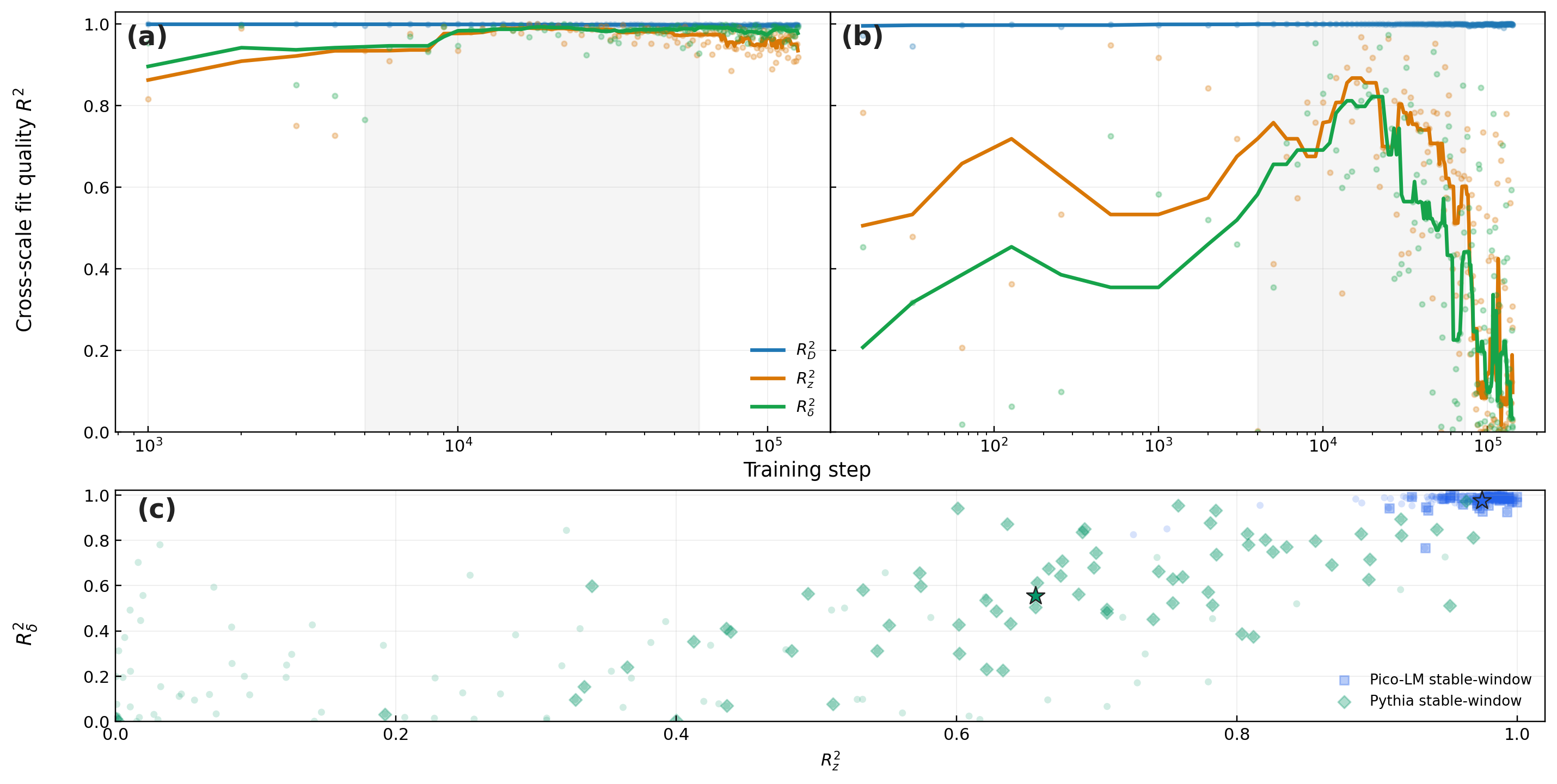}
    \caption[Stepwise compressibility contrast]{
    Stepwise compressibility contrast across the two model families.
    (a) and (b) show the time-resolved per-step cross-scale fit quality $R^2$ for the size,
    duration, and intensive-efficiency channels in Pico-LM and Pythia, respectively, with
    the stable windows shaded.
    (c) shows the $(R^2_z,R^2_\delta)$ organization map, where the upper-right corner
    corresponds to jointly compressible duration and efficiency channels.
    Faint background points show all steps passing the inclusion criteria, larger colored markers show stable-window
    steps, and stars show the corresponding stable-window means.
    Pico-LM (high $R^2_z$, high $R^2_\delta$) and Pythia (lower $R^2_z$, lower $R^2_\delta$)
    therefore occupy distinct regions of this organization map, identifying compressibility
    itself as a transport-organization observable rather than a fit-quality footnote.
    }
    \label{fig:compressibility_contrast}
\end{figure*}

Figure~\ref{fig:compressibility_contrast} shows that the cross-family regime
contrast is not only a contrast in exponent values, but also a contrast in
stepwise compressibility. By stepwise compressibility we mean whether a given
transport channel at a given training step is well described by a single
cross-scale power-law slope, quantified by the fit quality $R^2$. In Pico-LM,
the stable-window fit quality remains high across the fitted size, duration,
and intensive-efficiency channels shown in Fig.~\ref{fig:compressibility_contrast},
with mean $R^2_D\approx0.998$,
$R^2_z\approx0.975$, and $R^2_\delta\approx0.976$. In Pythia, the size backbone
remains equally clean ($R^2_D\approx1.000$), but the duration and efficiency
channels show much weaker one-slope compressibility
($R^2_z\approx0.656$, $R^2_\delta\approx0.555$).

This does not mean that the Pythia channels are absent. Rather, their
cross-scale structure is not well compressed by a single power-law slope at
each training step. For near-zero slopes such as the Pythia $z$ and $\delta$
channels, low $R^2$ should therefore be read as weak one-slope compressibility,
not as the disappearance of the underlying transport channel.

This distinction also constrains how internal transport can be compared with
external performance scaling. As quantified below, the strongest external
associations occur at the channel-observable level; by contrast, $D(t)$ acts
mainly as a shared size backbone, and exponent-level comparisons are further
limited by channel compressibility rather than yielding a significant
full-trajectory $D$--performance bridge in either family.
Compressibility should therefore be read as an observable of transport
organization rather than as a technical fit-quality footnote.

\subsection{Shared null baselines support cross-family comparison}
\begin{figure*}[!t]
    \centering
    \includegraphics[width=0.94\textwidth]{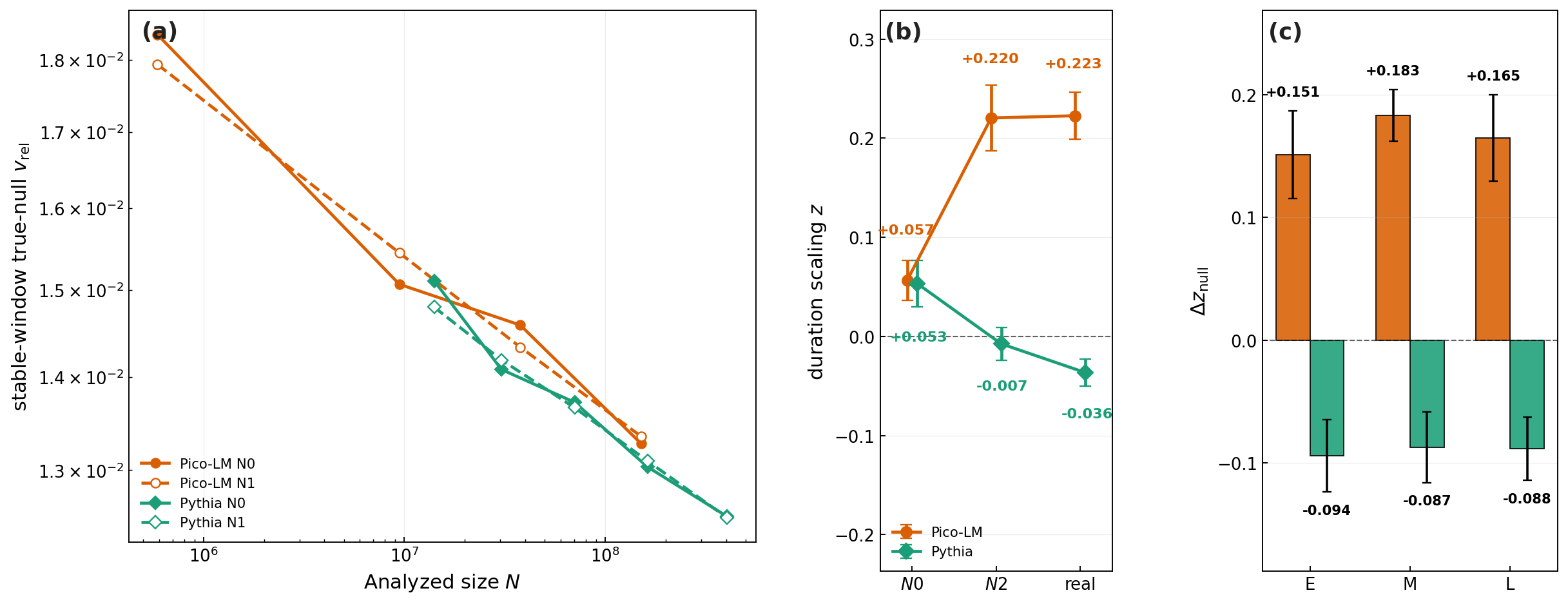}
    \caption[Shared null baselines across model families]{
    Shared null baselines and opposite duration departures across model families.
    (a) shows the stable-window $\Nzero/\None$ Gaussian null baselines of
    $v_{\mathrm{rel}}(N)$, which define a nearly shared null intensive-transport floor
    for Pico-LM and Pythia over the overlapping analyzed size range.
    (b) shows the duration-channel hierarchy from the Gaussian null baseline
    ($\Nzero$) to the distribution-preserving control ($\Ntwo$) and finally to the
    real curve, which adds the original value-to-node assignment correction.
    (c) shows the corresponding null-subtracted duration contrast
    $\Delta z_{\mathrm{null}}=z_{\mathrm{real}}-z_{\Nzero}$ across coarse early, mid,
    and late tertiles of the stable window; the same family colors as in (b)
    are used.
    The opposite signs in (c) show that the duration contrast is not a single-window
    artifact. Taken together, the three panels show that the families have closely
    matched null baselines in the displayed intensive-transport and duration
    readouts, while the real duration departures have opposite signs.
    }
    \label{fig:shared_null_skeleton}
\end{figure*}

Figure~\ref{fig:shared_null_skeleton} shows that the two families share closely matched
null baselines but exhibit opposite real departures in the duration channel.
Across overlapping analyzed size ranges, the $\Nzero/\None$ baselines for Pico and Pythia
lie on nearly the same null $v_{\mathrm{rel}}(N)$ floor [Fig.~\ref{fig:shared_null_skeleton}(a)].
The clearest overlap is Pico small versus Pythia 14M, where the $\Nzero$
stable-window mean $v_{\mathrm{rel}}$ values differ
by only about $0.3\%$. 

A parallel null-baseline agreement appears in the
duration channel: $z_{\Nzero}\approx+0.057$ for Pico and
$z_{\Nzero}\approx+0.053$ for Pythia
[Fig.~\ref{fig:shared_null_skeleton}(b)].
This makes it unlikely that the duration-channel contrast is driven by different
null calibrations. Instead, it arises from opposite real departures from these
nearly shared baselines. In the stable windows,
the null-subtracted duration contrast is
$\Delta z_{\mathrm{null}}=z_{\mathrm{real}}-z_{\Nzero}\approx+0.166\pm0.034$
for Pico and $\Delta z_{\mathrm{null}}\approx-0.090\pm0.028$ for Pythia
[Fig.~\ref{fig:shared_null_skeleton}(b)].
The sign does not flip under coarse early/mid/late partitioning of the
stable window [Fig.~\ref{fig:shared_null_skeleton}(c)].
This shared-baseline reading strengthens the comparison between a raw-gradient family
and a checkpoint-difference companion family without requiring the underlying microscopic
fields to be identical.
In this decomposition, the $\Nzero/\None$ baselines constitute the shared null
skeleton, $\Ntwo$ preserves the observed signed-value distribution under random
value-to-node assignment, and the real line adds the original value-to-node
assignment correction. In the duration channel, Pico lies close to the
distribution-preserving null ($z_{\Ntwo}\approx z_{\mathrm{real}}$). Pythia
retains a non-negligible assignment correction; in the full-run duration
decomposition, $\Delta z_{\mathrm{dist}}/\Delta z_{\mathrm{total}}\approx74.7\%$
and $\Delta z_{\mathrm{assign}}/\Delta z_{\mathrm{total}}\approx25.3\%$.

\subsection{Conservative performance bridge}
The preceding sections characterized the stable-window transport signatures of each 
family and their relation to randomized null baselines. 
We now ask whether the same internal transport observables are associated with
external performance over the available training trajectories, using the finite-scale
summaries $\beta_{\mathrm{PPL}}(t)$ and $\alpha_{\mathrm{acc}}(t)$ defined in
Methods.
For Pico-LM, the external trace is the per-checkpoint Paloma perplexity used to
form $\beta_{\mathrm{PPL}}(t)$ in Methods. For Pythia, we use the standard
seed-1234 zero-shot evaluation release~\cite{biderman2023pythia,eval-harness},
while the transport line uses seed-3 PolyPythias checkpoints~\cite{vanderwal2025polypythias};
because no matching seed-3 evaluation suite is publicly available, the Pythia
comparison is treated as companion evidence rather than as a seed-matched
predictive test.

\begin{figure*}[!t]
    \centering
    \includegraphics[width=0.96\textwidth,height=0.62\textheight,keepaspectratio]{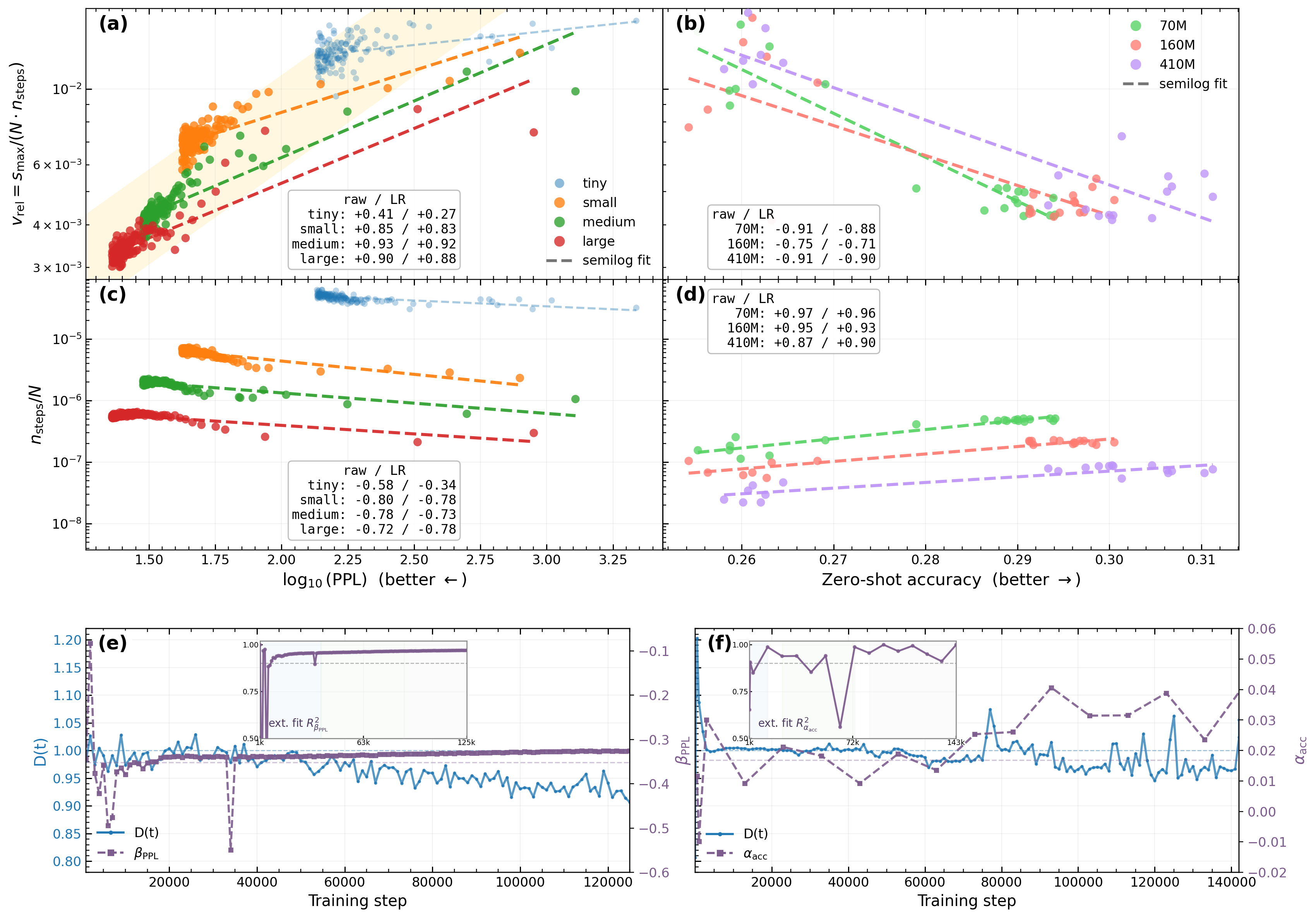}
    \caption[Conservative performance bridge]{
    Conservative performance bridge across model families.
    (a) and (b) show the association between external performance and
    $v_{\mathrm{rel}}$ for Pico-LM and Pythia, respectively; the pale band in
    (a) marks a visual pooled cross-scale envelope for the Pico points, not a
    confidence interval.
    (c) and (d) show the corresponding association for normalized duration
    $n_{\mathrm{steps}}/N$.
    (e) and (f) show the common size backbone $D(t)$ together with the
    external finite-scale exponents $\beta_{\mathrm{PPL}}(t)$ and
    $\alpha_{\mathrm{acc}}(t)$; the insets show the fit quality $R^2$ of the
    external exponents alone.
    Dashed lines are semilog trend guides and are not used to compute the
    reported correlations.
    Panel boxes report Pearson correlations as raw / learning-rate-partial
    values. 
    The panels therefore summarize the hierarchy used in the text:
    channel-level associations are clearest, while $D$ serves mainly as a shared
    size backbone rather than a strong full-trajectory performance correlate.
    }
    \label{fig:transport_channel_bridge}
\end{figure*}

Within this conservative setup, the strongest empirical associations are found
at the channel-observable level, especially for $v_{\mathrm{rel}}$ and
$n_{\mathrm{steps}}/N$. We pair this with a conservative
exponent-level check, asking whether internal transport exponents co-vary with
the external finite-scale exponents across training; as shown below, this check
does not yield a significant full-trajectory performance association in either family.
Figure~\ref{fig:transport_channel_bridge} summarizes both readings in one view:
the upper panels show the channel-level performance associations, while compact
lower panels provide the exponent-level check.
Because these correlations are computed along training trajectories, adjacent
checkpoints are temporally correlated; the reported $p$-values should therefore
be read as descriptive association statistics rather than independent-sample
significance tests.

In Pico-LM, the strongest empirical associations are channel-level. Within individual
models, $\log(\mathrm{PPL})$ is strongly positively correlated with
$v_{\mathrm{rel}}$ and strongly negatively correlated with normalized duration
$n_{\mathrm{steps}}/N$; the near-identical learning-rate-partial correlations
indicate that these associations are not explained by a linear dependence on
the family's reconstructed LR schedule alone.
Figure~\ref{fig:transport_channel_bridge}, panel~(a), further shows that the
same positive association appears across the cross-scale model ladder, so the
pooled Pico points suggest an approximate envelope in the
performance--$v_{\mathrm{rel}}$ plane. Panel~(c) is more structured: within a
fixed scale, lower PPL is associated with larger normalized duration, whereas
across scales the larger models occupy lower-PPL, lower-$n_{\mathrm{steps}}/N$
bands.

This association should not be read as a pointwise proxy for the instantaneous
learning signal. Even in the late stable windows
(Figs.~\ref{fig:vrel_hero}--\ref{fig:transport_analysis}),
$v_{\mathrm{rel}}$ still varies nontrivially within each Pico model, with
late-window coefficients of variation of approximately $5\%$--$8\%$, and the
medium model still shows detectable residual $v_{\mathrm{rel}}$ drift within the plateau window
($r\approx-0.48$, $p\approx0.006$, $n=31$). By contrast, the
exponent-level bridge is absent across the full trajectory:
$D(t)$ and $\beta_{\mathrm{PPL}}(t)$ are not significantly correlated 
($r\approx-0.10$, $p\approx0.26$, $n=125$; Fig.~\ref{fig:transport_channel_bridge},
panel~(e)). The inset in panel~(e) shows that this null result is not caused by
a collapse of the external finite-scale fit itself, because
$\beta_{\mathrm{PPL}}$ remains well fit after the early stage. We therefore use
$D(t)$ as the most conservative exponent-level check: it is the most stably
compressible cross-scale exponent, whereas $z(t)$ and $\delta(t)$ do not form
comparably clean or significant associations with the external exponents. Three
isolated Pico checkpoints (2k, 5k, and 34k) are retained for completeness but
do not alter this conclusion.

Pythia provides the complementary performance-association case. Here the strongest empirical
associations are again channel-level, but the matched comparison is more limited. Internal transport
reorganizes early: the move into a narrow band close to the $D=1$ baseline and a
narrow $v_{\mathrm{rel}}$ band occurs on the order of $10^2$--$10^3$ steps
(Figs.~\ref{fig:vrel_hero}--\ref{fig:transport_analysis}), whereas stable
external scale separation appears later, with $\alpha_{\mathrm{acc}}(t)$
becoming stably positive only after roughly $1.3\times10^4$ steps. Because the
evaluation checkpoints come from the seed-1234 standard release rather than the
seed-3 transport line, the matched comparison is limited to the 21-step overlap
between the two. On that overlap, cross-step $\alpha_{\mathrm{acc}}$ is
anticorrelated with mean $v_{\mathrm{rel}}$
($r\approx-0.62$, $p\approx0.003$) and positively correlated with normalized 
duration ($r\approx+0.67$, $p\approx8\times10^{-4}$). The corresponding
learning-rate-partial correlations remain close to these values
($-0.66$ for mean $v_{\mathrm{rel}}$ and $+0.77$ for normalized duration),
indicating that the channel-level association is not explained by a linear
dependence on the family's reconstructed LR schedule alone. This temporal separation between early internal reorganization and later
external accuracy-scale separation is at least qualitatively reminiscent of staged
circuit emergence during transformer training~\cite{olsson2022induction},
although the present observable is internal transport rather than circuit-level.

By contrast, the exponent-level association remains weak. The correlation between
$D(t)$ and $\alpha_{\mathrm{acc}}(t)$ is not significant
($r\approx-0.11$, $p\approx0.63$), and the duration/efficiency channels lose much
of their late-time stepwise compressibility. The inset in Fig.~\ref{fig:transport_channel_bridge}(f), shows the
corresponding limitation on the external side: $\alpha_{\mathrm{acc}}$ is fit
most cleanly in the late stage, but remains more variable through the mid-phase.
We therefore treat Fig.~\ref{fig:transport_channel_bridge}(f) as companion evidence rather than as a clean
exponent-level performance association. The isolated $\alpha_{\mathrm{acc}}$ point at step 63k is
retained for transparency but should not be overinterpreted.

The distinction between Pico and Pythia is therefore not whether any association
is visible, but which transport layer carries it. In both families the robust
shared layer is channel-level: $v_{\mathrm{rel}}$ and normalized duration track
external performance while $D$ acts as the common size backbone without itself
correlating significantly with external exponents across the available trajectory.
This stability is consistent with $s$ being the most integrated transport
observable, summing all activated units across a cascade event, which makes
the size law $s\!\sim\!N^{D}$ less sensitive to regime-specific decomposition
differences; that same aggregation may make $D$ less sensitive to the finer
per-step signals associated with external performance.
This performance association is therefore a consistency result about
transport-channel organization, not a derivation of neural scaling laws.

\section{Discussion}
The present evidence supports a decompositional view of finite-size transport
signatures in real language-model training. The central empirical contrast between the two families
is captured not by a single distinguishing exponent but by how the multi-channel
transport language partitions a shared near-unity $D$ backbone into different
duration and intensive-efficiency regimes. Pico and Pythia both 
admit a clean cascade-size backbone ($D\approx1$, with high stepwise $R^2_D$
across the retained trajectory); family differences emerge less in whether this backbone 
exists and more in how it is decomposed into duration ($z$) and intensive 
efficiency ($\delta$) channels: Pico converts scale into longer cascades and 
lower per-parameter efficiency, whereas Pythia keeps duration nearly scale-flat 
and concentrates into a narrow efficiency band. This distinction is visible directly in $v_{\mathrm{rel}}$, 
summarized compactly by the five-quantity closure, and reinforced by both the 
null-decomposition evidence and the compressibility contrast. This reading is also compatible
with broader work on critical signal propagation and training-phase structure in
deep networks, including order-to-chaos analyses~\cite{poole2016expressivity,schoenholz2017deep},
edge-of-stability dynamics~\cite{cohen2021edge}, and large-learning-rate phase
behavior~\cite{lewkowycz2020catapult}.

This perspective also clarifies the relation between the present analysis and
the prior toy-grokking applications of the TDU-OFC cascade response probe. In modular addition
and XOR, the effective cascade dimension $D(t)$ exhibits a localized crossing
of the $D{=}1$ baseline that coincides with the generalization transition, with
a task-dependent crossing direction, and then, most clearly in modular addition,
remains near $D{=}1$ after the grokking
transition~\cite{wang2025tduofc_a,wang2025tduofc_b}. In the
real language-model pretraining runs analyzed here, the same near-$D{=}1$ residence is
reached without an intermediate sharp behavioral transition: in Pythia, after
a brief startup descent from $D{\approx}1.2$, which resembles toy ModAdd in
its downward approach to the $D{=}1$ band but not in its anchoring to a sharp
behavioral transition; in Pico-LM, essentially from the beginning
of the analyzed window, with only a weak systematic late-time deviation. The
two settings therefore share a near-unity post-transient $D(t)$ band rather
than an identical crossing event, and differ mainly in how that band is
reached: a sharp grokking transition in the toy benchmarks, a brief startup descent
in Pythia, and near-immediate residence in Pico-LM over the analyzed window. 

The toy/LLM contrast is therefore
not simply that the toy crossing ``broadens'' into a regime, but that $D(t)$ changes
role: from a transition-localized marker in controlled grokking to a
near-unity cascade-size backbone for real language-model pretraining. This shared
post-transient band is calibrated by the prior synthetic
Gaussian-gradient controls, which identified $D{=}1$ as the
TDU-OFC cascade response probe's Gaussian-null baseline~\cite{wang2025tduofc_a}. The
qualitative size-channel null reading reported with the family-specific
evidence further suggests that this residence is not simply a Gaussian-noise
floor. In Pythia, empirical value distributions appear to lift $D$ above the
baseline, while assignment structure offsets part of that lift and leaves the
real trajectory near $D{=}1$. This pattern is consistent with a partial
cancellation between distribution and assignment contributions rather than with
a complete absence of structure. We mark this as
qualitative because the size-channel real--null gap is small; the formally
quantified cross-family null decomposition is reported in the duration and
intensive-transport channels (Fig.~\ref{fig:shared_null_skeleton}), where the
real--null gap is larger and supports a quantitative distribution/assignment
split.

$D$ itself remains the most stably extractable cross-scale exponent in both
settings, with stable-window $R^2_D \approx 1.0$ in both Pico-LM and Pythia
(Fig.~\ref{fig:compressibility_contrast}), so the five-quantity framework
introduced here is built around $D$ rather than as a substitute for it. The
exact closure relations $\beta=D-z$ and $\delta=D-z-1$ constrain the additional
channels algebraically through $D$ and $z$. The extension from a single size
exponent to multi-channel finite-size gradient transport observables therefore
enriches the single-exponent analysis with finer structure in the duration,
intensive-efficiency, and compressibility channels. At the same time, it
preserves $D$ as the shared near-unity backbone and as the most directly
comparable observable across the toy-grokking and real language-model
pretraining studies.

This framework also sharpens what may and may not be claimed. Pico should be 
read as showing only a limited approach toward the $D=1$ baseline plus a persistent 
efficiency-suppression regime, not as a stable above-baseline phase. Pythia 
should be read as occupying a narrow close-to-unity band during the stable 
window with weak positive efficiency scale dependence, not as a proven fixed point and not 
as a full dynamic data collapse. The shared-null-skeleton result strengthens 
cross-family comparability by showing that the null floor is nearly family-independent, 
yet the representation difference remains an important caveat, so strong quantitative claims stay 
within-family while cross-family claims remain regime-level.

The association with external scaling is correspondingly conservative and channel-level. 
External performance does not appear as a pointwise readout of internal transport. 
Instead, the dominant shared layer is channel-specific: better external
performance is associated with lower $v_{\mathrm{rel}}$ and higher normalized
duration, while $D$ acts as a common size backbone rather than the dominant
full-trajectory correlate. Across the full training 
trajectory, $D(t)$ does not correlate significantly with external exponents in 
either family, so no 
exponent-level performance-association claim is warranted. This is consistent with a transport
picture accompanying stable external scaling, but it does not derive neural
scaling laws from first principles.
The present evidence remains observational and finite-scale: it does not
isolate causal effects of optimizer, schedule, architecture, or data order, and
the Pico/Pythia contrast changes several axes at once.

The framework yields limited, falsifiable regime predictions. If a larger
raw-gradient family remains Pico-like, then positive duration scaling and
negative intensive transport scaling should persist together with high stepwise
compressibility in $z$ and $\delta$. If a family remains Pythia-like, then $D$
should stay close to the $D=1$ baseline while the duration/efficiency channels
enter a weakly ordered narrow band whose stepwise dynamics are no longer compactly
summarized by single power-law exponents, even as the raw channels
$v_{\mathrm{rel}}$ and $n_{\mathrm{steps}}/N$ remain functionally real and
empirically associated with performance. A more tentative hypothesis concerns temporal pacing:
in the present data, Pico shows a stronger observed alignment between $D(t)$
and the learning-rate schedule than Pythia does. Whether that contrast persists
in larger families can be tested directly with denser checkpoints, alternate
optimizers, and matched evaluation traces. These are regime-level hypotheses
rather than universal laws.
This perspective is also consistent with mechanistic studies of grokking, where
continuous internal progress measures precede an apparently sharp behavioral
transition~\cite{nanda2023grokking}, and with induction-head analyses linking
the emergence of a specific internal circuit to a sharp increase in in-context
learning ability~\cite{olsson2022induction}.

The broader path forward is therefore not to search for one universal crossing
pattern. It is to extend the same transport language across more model families,
optimizers, and schedules, and to treat compressibility itself as an observable
of internal organization. A further extension is to apply the same diagnostic
protocol to larger mainstream language-model pretraining runs. For instrumented
training runs, gradient or update snapshots could support offline,
noninterventional transport analysis along the training trajectory; when only
saved checkpoints are available, checkpoint-difference fields could support
retrospective comparative analysis, provided that tensor families, checkpoint
density, optimizer metadata, and field representations are controlled
consistently. In this sense, the present work contributes less a one-off
classification and more a reusable non-equilibrium statistical mechanics framework for measuring
how real language-model training runs reorganize, and how internal transport
responses change, during training. The TDU-OFC construction used here should
therefore be read as one concrete thresholded response probe for gradient and
update fields, not the only possible member of a broader family of
event-triggered or history-dependent response probes.

\section{Conclusion}
We extend the single-exponent $D(t)$ analysis into a finite-size gradient
transport framework for real LLM training. The same five-quantity closure
applies to Pico-LM and Pythia, and both families share a near-unity size
backbone. They decompose that backbone into distinct regimes: Pico-LM shows
duration-supported transport with scale-suppressed intensive efficiency,
whereas Pythia remains in a narrow band close to the $D=1$ baseline with weaker
one-slope compressibility in the duration and efficiency channels.

Together with the shared null skeleton and the conservative channel-level
association with external performance, these results support a reusable
transport framework, grounded in non-equilibrium statistical mechanics, for comparing real
training regimes without claiming a universal fixed point or a first-principles
derivation of neural scaling laws.
In this sense, the present contribution is a set of SOC-inspired transport
\emph{observables} that resolve real-pretraining organization, rather than a
claim that LLM training itself realizes SOC dynamics.

The most informative next test would be a controlled additional family, or set
of matched runs, that separates optimizer choice (for example, Adam versus
momentum-only training) from architecture and field representation. Matched
checkpoints and a common evaluation suite would allow
the same five-quantity closure and compressibility observable to distinguish
whether the Pico/Pythia regime contrast arises primarily from the
raw-gradient-versus-update-field distinction or from the model-family
distinction. Such a test would turn the present regime contrast from a
comparative observation into a sharper diagnostic of which ingredients control
finite-size transport in real training.

\section*{Data and Code Availability}
The pretrained model weights analyzed in this work are publicly available:
Pico-LM checkpoints from the \texttt{pico-lm/pico-decoder-\{tiny,small,medium,large\}}
repositories on the Hugging Face Hub~\cite{pico2025}, and Pythia checkpoints from the
\texttt{EleutherAI/pythia-\{14m,31m,70m,160m,410m\}} repositories~\cite{biderman2023pythia}.
The analysis code, the post-processed per-step temporal-dynamics JSONs for each
analyzed trajectory, and the summary JSONs supporting the numerical values
reported in the figures will be made available in a public repository upon
publication.
The release will use \texttt{CritScope} as the software pipeline name, while
TDU-OFC denotes the cascade response probe within the finite-size
gradient-transport measurement framework.
It will provide a reusable pipeline for finite-size gradient transport analysis, including
standardized temporal-dynamics and figure-summary JSON formats, together with
dependency and provenance metadata needed to reproduce the figures from the
released summary files.

% ============================================================================
% ACKNOWLEDGMENTS
% ============================================================================

\begin{acknowledgments}
% NOTE: Please fill in before submission
Ping Wang acknowledges support from the National Key R$\&$D Program of China 
(grant Nos. 2024YFA1611701, 2024YFA1611700).

\end{acknowledgments}

\bibliographystyle{apsrev4-2}
\bibliography{refs}

\end{document}